# Text Guide: Improving the quality of long text classification by a text selection method based on feature importance

**Krzysztof Fiok[1], Waldemar Karwowski[1], Edgar Gutierrez[1,2,\*], Mohammad Reza Davahli[1], Maciej Wilamowski[3], Tareq Ahram[1], Awad Al-Juaid[4], and Jozef Zurada[5]**

[1] Department of Industrial Engineering and Management Systems, University of Central Florida, Orlando, FL 32816, USA; fiok@ucf.edu (K.F.); wkar@ucf.edu (W.K.); MohammadReza.Davahli@ucf.edu (M.R); tahram@ucf.edu (T.A.)
[2] Center for Transportation and Logistics, Massachusetts Institute of Technology, Cambridge, MA 02139, USA (E.G)
[3] Faculty of Economic Sciences, University of Warsaw, Warsaw, Poland; macwilam@gmail.com (M.W.)
[4] Department of Industrial Engineering, College of Engineering, Taif University, Taif 21944, Saudi Arabia; amjuaid@tu.edu.sa (A.A.)
[5] Business School, University of Louisville, Louisville, KY 40292, USA; jmzura01@louisville.edu (J.Z.)

Corresponding author: Edgar Gutierrez (e-mail: edgargutierrezfranco@gmail.com )

This study was supported in part by a research grant from the Office of Naval Research (N000141812559) and was performed at the University of Central Florida, Orlando, Florida. Additional support was provided by Taif University Researchers Supporting Project, number TURSP- 2020/229, Taif University, Taif, Saudi Arabia.

**ABSTRACT** The performance of text classification methods has improved greatly over the last decade for text instances of less than 512 tokens. This limit has been adopted by most state-of-the-research transformer models due to the high computational cost of analyzing longer text instances. To mitigate this problem and to improve classification for longer texts, researchers have sought to resolve the underlying causes of the computational cost and have proposed optimizations for the attention mechanism, which is the key element of every transformer model. In our study, we are not pursuing the ultimate goal of long text classification, i.e., the ability to analyze entire text instances at one time while preserving high performance at a reasonable computational cost. Instead, we propose a text truncation method called Text Guide, in which the original text length is reduced to a predefined limit in a manner that improves performance over naive and semi-naive approaches while preserving low computational costs. Text Guide benefits from the concept of feature importance, a notion from the explainable artificial intelligence domain. We demonstrate that Text Guide can be used to improve the performance of recent language models specifically designed for long text classification, such as Longformer. Moreover, we discovered that parameter optimization is the key to Text Guide performance and must be conducted before the method is deployed. Future experiments may reveal additional benefits provided by this new method.

## I. INTRODUCTION

During the period of 2010–2020, the performance of automated text classification improved significantly owing to the concept of creating vector representations for textual entities, often referred to as "embeddings." The work by [1] demonstrated a method for creating embeddings in an automated manner that outperformed language models developed with the use of human expert knowledge from the linguistic domain, i.e., lexicon-based approaches. Later, other researchers developed more sophisticated methods for obtaining embeddings of text entities. Experiments with various deep neural networks (DNNs) such as recurrent neural networks, convolutional neural networks, and, more recently, graph neural networks [2] have led to the proposal of increasingly more complex language models capable of outputting vector representations. However, since [3] demonstrated the benefits of the "attention" mechanism in deep learning models used for natural language processing (NLP), the so-called transformer models, which use various attention variants, have demonstrated state-of-the-art performance in numerous text analysis tasks. Based on these considerations, when the highest possible performance of a language model is required and short text instances are analyzed, the model selection process is generally focused on appropriate transformer models.

However, for researchers and machine learning practitioners analyzing long textual instances, the model selection process extends beyond transformer models. In this context, the decision process is more complicated; for instance, classical text classification approaches are considered. More traditional language modeling methods focus on counting occurrences of words stored in human-curated lexicons, such as Linguistic Inquiry and Word Count (LIWC) [4] or Sentiment Analysis and Cognition Engine (SÉANCE) [5], or all tokens and possibly ngrams in the text instance, as in the bag of words (BoW) or term frequency methods; thus, the computational costs associated with these methods grow slowly with increasing text instance length. Therefore, these methods can be deployed to gather information from the whole text instance regardless of its length. Unfortunately, for DNNs, creating vector representations of text is computationally expensive, and the growth rate of computation costs can be significant. To mitigate the rapid increase in computational cost, most transformer models have a limit of 512 for the number of tokens that they analyze (NTA) at one time.

## A. SOLUTIONS FOR LONG TEXT CLASSIFICATION

Many recent works have sought to alleviate this constraint and to find a general solution for applying more sophisticated DNN language models to longer text instances. Table 1 lists possible approaches that have already been explored. The computational costs were subjectively rated based on approximations from experiments with methods # 1, 3, and 6. Precise estimation of the computational cost of other methods was impossible as they were published without open access to their codes.

TABLE I.
POSSIBLE APPROACHES TO ANALYZING TEXT INSTANCES LONGER THAN THE NTA LIMIT. THE COMPUTATIONAL COST WAS SUBJECTIVELY RATED FROM 1 (LOWEST) TO 5 (HIGHEST). THE LOSS OF INFORMATION IS UNDERSTOOD AS THE INABILITY TO ANALYZE THE WHOLE TEXT INSTANCE.

| No. | Name of Approach | Description | Loss of Information | Computational Cost | Reference |
|---|---|---|---|---|---|
| 1 | Naive head-only | Truncation method. Use first NTA tokens from the text instance. | TRUE | 1 | Used in numerous frameworks, for example [6] |
| 2 | Tail-only | Truncation method. Use last NTA tokens from the text instance. | TRUE | 1 | [7] |
| 3 | Part1 head + Part2 tail | Truncation method. Use Part1*NTA tokens from the beginning of the text instance and add Part2*NTA tokens from the end of the text instance. Assume Part1 + Part2 = 1. The authors proposed Part1 = 0.2 and Part2 = 0.8. | TRUE | 1 | [7] |
| 4 | Divide, embed, join | 1) Divide the whole text instance into parts that fit the NTA limit, 2) deploy the text embedding model on each of the parts, and 3) join information from each part to create instance-level representation. Various joining approaches are available: a) simple averaging of part-embeddings, b) DNN training on top of part embeddings. | FALSE | 5 | [8] |
| 5 | Divide, embed, join variants | Optimized variants of the "divide, embed, join" approach. For example, the standard transformer model architecture can be extended by injecting recurrent elements, enabling one to capture long-range dependencies. | FALSE | 5 | [9] |
| 6 | Increased NTA for transformer models | A collection of approaches to modify the standard attention mechanism, with the common aim of lowering the computational cost and increasing the NTA limit for transformer models. | True if instance length still exceeds the NTA | 3-4 | Various authors, for example [10] [11] [12] |

| 7 | Sentence selection method based on encoder–decoder | A method can select "sentences more relevant" for text analysis from the whole instance based on encoder-decoder architecture. | True, but lesser | 3 | [13] |
| 8 | MemRecall | A method can select blocks of text relevant to the NLP task according to a second "judge" transformer model. | True, but lesser | 3 | [14] |

As shown in Table 1, there is currently no perfect approach. Approaches 1–3 are naive or semi-naive truncation methods that lose information but have a minimum computational cost. Approaches 4 and 5 do not lose information, but are computationally expensive, as the base language model must be applied several times and additional steps are needed in order to merge the partially extracted information into a single instance-level representation. For the group of methods listed as Approach 6 in Table 1 ("Increased NTA for transformer models"), it is difficult to increase the relatively low length limit of standard transformer models in a manner that does not drastically increase the associated computational cost. This research direction has been extensively explored in recent studies. The rationale for these efforts seems obvious; namely, if the NTA can be sufficiently increased while preserving a low computational cost, then there is no need for loss of information, and analyzing the whole text instance with a single model has obvious advantages over deploying a model numerous times over separate parts and merging the obtained information. In the transformer model domain, researchers have proposed increasing the NTA from 512 to 4096 tokens and have introduced numerous modifications to the core attention mechanism, with a notable example of a "sparse attention mechanism" by [15]. Extensive reviews and comparisons of numerous methods in this field have been previously presented in [16] and [17].

Finally, approaches 7–8 in Table 1 allow a loss of information, but aim to select parts of the original text instance in an intelligent manner to preserve the most relevant elements. Due to their deep learning nature, these methods are complex, and their computational cost is substantially higher than that of the truncation methods (Approaches 1–3).

To summarize, researchers seeking to achieve better long text classification are focused on the tradeoff between loss of information and computational cost.

### B. CONTRIBUTION OF THIS STUDY

If researchers seeking to address a new long text classification task were to conduct a brief review of Table 1, they could conclude either that the whole task can be kept simple with naive and semi-naive truncation (Approaches 1–3) and standard transformer models or that complex deep learning approaches can be applied to analyze whole text instances or iteratively analyze select parts of the original text instance in a complex manner.

The present study contributes to this aim by demonstrating a new approach called Text Guide, which proposes an easy-to-apply yet an intelligent solution for selecting parts of the original text instance. This approach was inspired by the methods of [13] and [14], which focused on intelligent sentence selection. Text Guide falls into the same category as sentence selection methods; however, Text Guide has a simpler and less computationally expensive design. As such, Text Guide is not as simple as the previously discussed truncation methods, but not as complex as known deep learning methods.

Text Guide borrows from the domain of explainable artificial intelligence (XAI). Specifically, Text Guide utilizes features and their importance, which are output from a pre-trained machine learning classifier to select relevant text fragments.

The main advantages of Text Guide are as follows:
1) Text Guide is a model-agnostic approach, meaning it does not interfere with the later selected language model. For example, any available deep learning model can benefit from the Text Guide.
2) Text Guide is computationally inexpensive.
3) Text Guide provides a tangible improvement of final classification performance over baseline truncation methods.

To demonstrate its performance, we deployed Text Guide on three datasets, utilizing two recent transformer models with different NTAs, and explored how the choice of Text Guide parameters affects the final classification performance. To provide easy application of our method by the research community, we have released our code and a data sample and example script that allows using Text Guide [18].

## II. METHODS

### A. TEXT GUIDE METHOD

The proposed approach for classifying long text data can be divided into three phases, as presented in Figure 1.

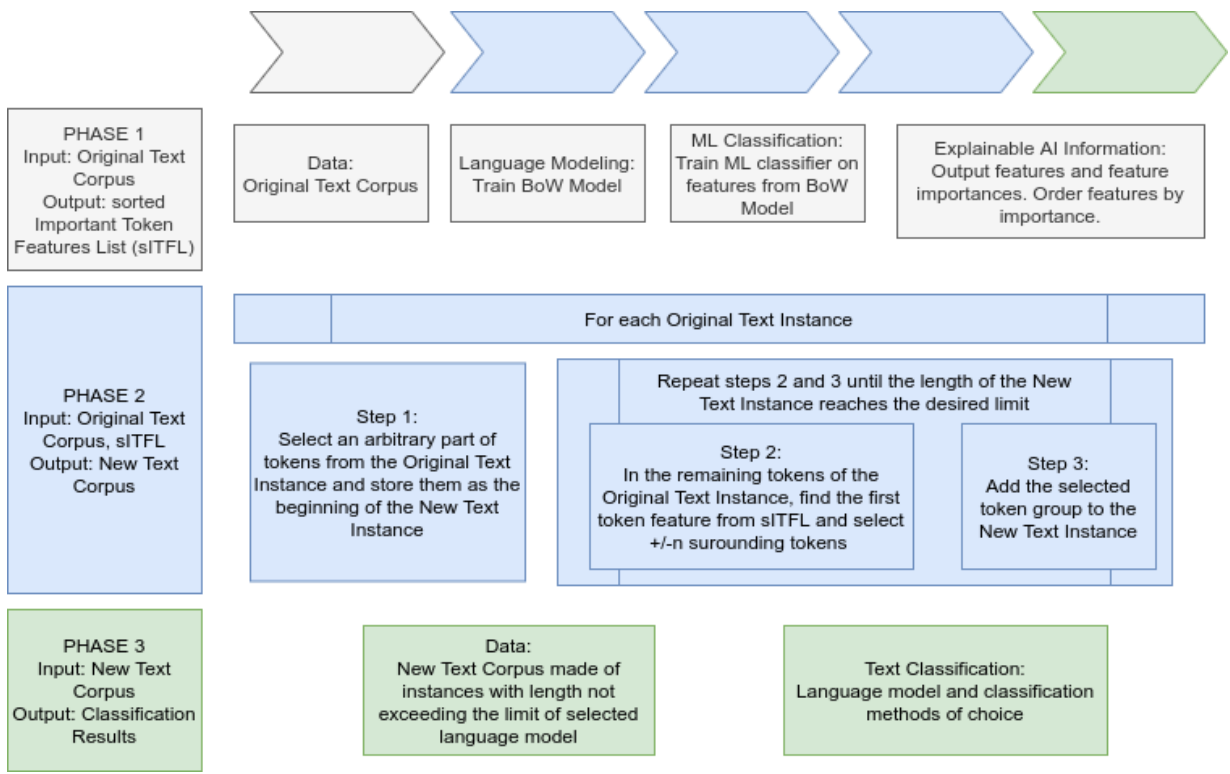

**FIGURE 1.** Three phases of analysis in the proposed Text Guide method.

1) PHASE 1

The goal of the first phase is to obtain a list of tokens sorted by their relevance to the classification task in question. To create such a list, we perform the following tasks:

  a) Train a classic BoW model on the original text corpus. The output from the BoW model is a list of token features and the number of occurrences per text instance. To improve model performance and reduce overfitting, we select only N features characterized by the highest mutual information.
  b) Train a machine learning classifier with the use of N features from the previous step. In our experiments, we employed a gradient boosting classifier,
  c) Employ a selected XAI method to output model-level feature importances, after the classifier is trained. In our experiments, we obtained feature importances identified by the gradient boosting classifier.
  d) Sort the token features used by the machine learning classifier based on the feature importance, starting from the highest value.

After step 4, the first phase ends, and the sorted list of N important token features (sITFL) is input to the second phase.

2) PHASE 2

The second phase depicted in Figure 1 is the core of *Text Guide*. In this phase, we iteratively process each original text instance that exceeds the desired length limit defined as the NTA and extract parts of the original text in order to create a new text instance that fits the NTA. The second phase is performed via the following steps:

  a) Select an arbitrary portion of tokens from the original text instance, store this portion as the beginning of the new text instance, and delete the portion from the original text instance. The arbitrary portion of tokens can be extracted from the original text instance like the truncation methods (Approaches 1–3) presented in Table 1. However, with these previous methods, the whole new text instance is simply truncated from the original text instance. In contrast, we create only a small fraction of the new text instance in this manner. For example, if Approach 3 from Table 1 is adopted, we define Part1 = 0.2 and Part2 = 0.1. Specifically, we aim to keep Part1 + Part2 << 1, so that the remaining Part3 = 1 - (Part1 + Part2) maintains a reasonable value.
  b) To fill the new text instance with the remaining Part3*NTA tokens, iterate over the sITFL

originating from phase 1. Starting with the token n previously identified as the most important, we search for this token in the original text instance. If this token is not found, we select the next most important token n+1, and search again. However, if the selected token exists in the original text instance, we select this token together with its token neighbors (TNs) located before and after. This sub-step is visualized in figure 3, representing how token groups are defined in a fragment of an exemplary original text instance.

c) Add the resulting token group of TN tokens+selected important token+TN tokens to the new text instance. Depending on the selected number of TNs, the added token groups will provide more or less context to the selected important token.

d) Repeat steps 2 and 3 until the whole new text instance is filled and its length reaches the NTA. A visualization illustrating how a new text instance is formed according to the described procedure is presented in Figure 2.

e) Repeat steps 1–4 for all instances in the original text corpus in order to create a new text corpus with all new text instances meeting the NTA requirement.

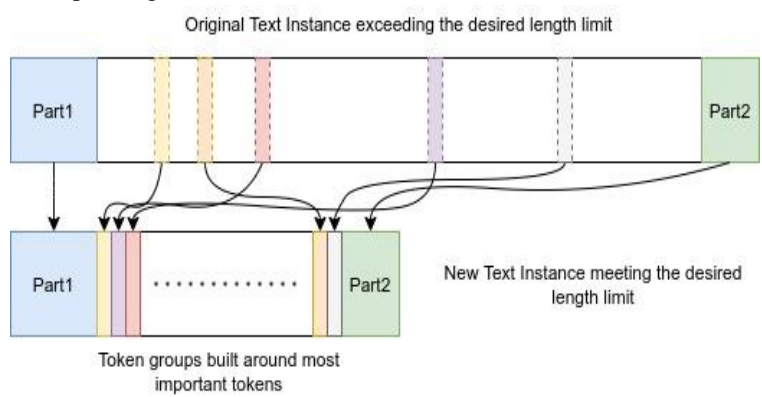

**FIGURE 2.** Visualization of an example of a new text instance formed according to the procedure described as phase 2 of the proposed approach. Part1 and Part2 are arbitrarily selected from the original text instance, whereas token groups are built around the important tokens found in the original text instance and copied to the new text instance.

State-of-the-art Natural Language Processing for Jax, Pytorch and TensorFlow 😊 Transformers (formerly known as *pytorch-transformers* and *pytorch-pretrained-bert*) provides general-purpose architectures (BERT, GPT-2, RoBERTa, XLM, DistilBert, XLNet…) for Natural Language Understanding (NLU) and Natural Language Generation (NLG) with over 32+ pretrained models in 100+ languages and deep interoperability between Jax, PyTorch and TensorFlow.

**FIGURE 3.** Visualization of sub-step b) from Phase 2 of Text Guide on a fragment of an exemplary text instance (text copied from [19]). Assuming TN = 2 and that sITFL was a list of tokens comprising of 'processing', 'architectures', 'deep', and other tokens not existing in the analyzed text fragment, Text Guide extracts three token groups from the example text. The located important tokens are colored in green, blue and red, whereas neighbor tokens are colored in yellow. The resulting token groups that will be used in the new text instance created by Text Guide are "natural language processing for jax", "general purpose architectures bert gpt", and "languages and deep interoperability between".

3) PHASE 3

When the third phase of *Text Guide* is reached, the problem of long text classification is reduced to a classification task for a "standard-length" text using the selected language model. Therefore, it is possible to benefit from standard modeling; for example, if an NTA of 510 is set, all mainstream transformer models with an NTA of 512 can be utilized.

*B. Explaining Text Guide by example*

Assume the original text instance has a length of 700 tokens and that the deep learning language model selected for analysis accepts only texts of up to 300 tokens long. We want to truncate the original text instance and select Text Guide as the method. Since using Text Guide requires an sITFL, let us assume that we have defined it already in Phase 1 on our own as a list of 100 important tokens. Therefore, we begin Phase 2 from sub-step a), i.e., select an arbitrary portion of tokens from the original text instance. We choose parameter values Part1=0.1 and Part2=0.05 which leads to taking 700*0.1=70 tokens from the beginning and 700*0.05=35 tokens from the end of the original text instance. These parts are used to form the beginning of the new text instance. After this step, the new text instance has a length of 70+35=105 tokens, which is below the desired NTA of 300 tokens. Therefore, we continue to sub-step b) and aim to fill the remaining 300-105=195 tokens by groups of tokens defined according to *Text Guide*. Let's assume we want to create short token groups as in figure 3 and define TN=2. We begin sub-step 2) by taking the first token from sITFL and discover that it does not exist in the currently analyzed original text instance. Thus, we continue and take the second token from sITFL and discover that it exists in the original text instance. Therefore, we can define the first token group as in figure 3. Because we take the important token and TN=2 tokens from both sides of the important token, the extracted token group is of length 2+1+2=5 tokens. After this step, the new text instance is of length 105+5=110 tokens, so there is still free space for 300-110=190 tokens in the new text instance. Therefore, we look for the third, fourth, fifth… essential tokens in our sITFL and extract token groups until the space in the new text instance is exhausted. We discover that we filled the whole 300 token lengths of the new text instance after the 74th important token from sITFL. Finally, the less important tokens are neglected, phase 2 is finished, and the new text instance is formed by *Text Guide*. The selected deep learning language model can then be deployed.

### C. TEXT GUIDE PARAMETER OPTIMIZATION

*Text Guide* can provide various results depending on the parameters adopted in phase 2, such as Part1, Part2, and TN. To assess the influence of these parameters on the performance of *Text Guide*, we conducted numerous trials with the following values: Part1: {0.1, 0.2, 0.3, 0.4, 0.5}; Part2: {0, 0.05, 0.1, 0.15}; TN: {1, 2, 3, 4, 5, 6, 7, 8, 9, 10}.

Adoption of these values allowed us to test various situations, for example, 1) when a large part of the new text instance is created from the beginning of the original text instance, 2) when a small part of the new text instance is created from the beginning of the original text instance, 3) when the TN number is small and little context is provided for the important feature token, but the more important feature tokens fit in the adopted NTA limit, 4) a situation opposite to situation 3, 5) when Part2 is equal to 0, i.e., the ending part of the original text instance is neglected, 6) when Part2 is equal to 0.15, i.e., a considerable portion of the ending part of the original text instance is retained.

### D. DATA SETS USED FOR BENCHMARKING

Various data sets have been used to demonstrate the performance of the methods presented in Table 1. Interestingly, some of these methods included very small numbers of long text instances; for example, [14] experimented on 20 Newsgroups [20] and reported that only 15% of the approximately 19 000 text instances available in that data set exceeded the most common 512 token limit. We believe it is difficult to determine the difference in quality between proposed methods when a mix of text instances that do and do not exceed the length limit are analyzed. To overcome this issue, we benchmarked *Text Guide* on the open DMOZ [21] data set, with text instances formed from parsed websites categorized into over 500 categories. The full DMOZ data set consists of 345 003 text instances, with the majority being shorter than 512 tokens and a significant portion exceeding that limit. For the purpose of our analysis, we selected only instances exceeding the 510 limit and formed three data sets, as presented in Table 2. In Figure 4, we present cumulative distribution functions of text instance lengths in the adopted data sets.

TABLE II
DESCRIPTION OF UTILIZED BENCHMARKING DATA SETS.

| Data set | Number of Instances | Length of Instances | Adopted NTA |
|---|---|---|---|
| DMOZ 510-1500 | 18 732 | Between 510 and 1500 tokens | 510 |
| DMOZ 1500+ | 8 379 | Over 1500 tokens | 510 |
| DMOZ 4090+ | 2 737 | Over 4090 tokens | 4090 |

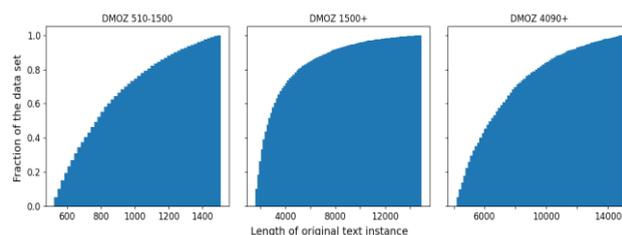

FIGURE 4. Cumulative distribution function of text instance lengths in the adopted data sets. For the sake of clarity, for DMOZ 1500+ and DMOZ 4090+ text instances of length exceeding 15000 tokens were not visualized as they constitute less than 1% of the data set.

The rationale for selecting 510, rather than 512, tokens is that transformer models add two special tokens at the end and beginning of each text instance. For all data sets, we extracted the 30 most common categories from the full DMOZ data set to maintain a reasonable number of data instances in each category, and we deleted instances with duplicated text. The rationale for the definitions of the first two proposed data sets was to determine the performance of *Text Guide* for situations in which the adopted NTA limit is exceeded by the original text instance length to various extent. 1) In DMOZ 510-1500, text instances exceed the adopted NTA limit of 510 to a moderate extent. 2) In DMOZ 1500+, text instances exceed the adopted NTA limit of 510 tokens to a high extent. We created the DMOZ 4090+ data set to demonstrate that our method also provides good results when text instances are very long and the adopted NTA of 4090 is high but still lower than the actual length of the original text instances. DMOZ 4090+ was also introduced to demonstrate that the use of *Text Guide* can be beneficial for text instances that exceed the NTA limit of 510 to an extreme extent.

### E. LANGUAGE MODELS USED FOR BENCHMARKING

#### 1) TEXT GUIDE PARAMETER OPTIMIZATION FOR AN NTA OF 510

In our study, we focused on transformer models used for text classification, as they usually provide state-of-the-art performance. In our benchmarks on the DMOZ 510-1500 and DMOZ 1500+ data sets, we adopted a small transformer model called Albert [22], which has the standard NTA limit of 512. To avoid bias stemming from fine-tuning or other training procedures that can be performed for language models, we utilized the pre-trained model version from the transformer library [19].

#### 2) TEXT GUIDE PARAMETER OPTIMIZATION FOR AN NTA OF 4090

The selection of data sets from Table 2 allowed us to also demonstrate that *Text Guide* provides good results for NTA values way greater than 510. In particular, using the DMOZ 4090+ data set, we demonstrate that recent transformer models, such as Longformer c, with an NTA of 4096 can also benefit from *Text Guide* if the analyzed text instances exceed the token limit. In this setup, we also used the pre-trained model version from the transformer library.

#### 3) DEMONSTRATION OF TEXT GUIDE AND FINE-TUNED MODELS

Finally, to demonstrate that *Text Guide* provides quality improvement for models trained on the downstream task, we present results obtained by Albert and Longformer [10] models fine-tuned in a traditional manner, i.e., according to the naive methods (Approaches 1 and 3) presented in Table 1, in comparison to models fine-tuned on data obtained according to our best-performing *Text Guide* variant. Unless stated otherwise, the parameters for fine tuning of the selected models were adopted according to Table 3. To enrich the comparison, we also present results obtained by the state-of-the-research RoBERTa base [23] model fine-tuned on data instances prepared by Text Guide and a plain BoW model, precisely the ones used earlier for providing sITFL to Text Guide. Non-disclosed parameters were adopted as the default values proposed in the Flair (version 0.6 post 1) Python package.

### F. MACHINE LEARNING REGIME

Due to the existence of 30 categories and the imbalanced nature of the analyzed data sets, all of our experiments were executed using five-fold stratified cross validation. Following the recommendation of [24], we employed the Matthews correlation coefficient (MCC) as a performance metric. Each utilized transformer language model was applied in a manner such that it outputs a vector representation of the text instance via the classification token, and the provided text embedding was further analyzed by a gradient boosting classifier.

TABLE III
PARAMETERS ADOPTED FOR FINE TUNING OF SELECTED MODELS.

| | Model | | |
|---|---|---|---|
| Parameter | Albert Base-v2 | Longformer Base | RoBERTa Base |
| Initial Learning Rate | 3.00E-05 | 3.00E-06 | 3.00E-06 |
| Batch Size | 24 | 2 | 4 |
| Minimal Learning Rate | 3.00E-06 | 3.00E-07 | 3.00E-07 |
| Maximum Epochs | 6 | 6 | 6 |

The adopted batch sizes are a result of model size and the limitations of our 24 GB GPU ram computing machine. the learning rates had to be independently adjusted for both models to achieve convergence.

For experiments in which the transformer language model was fine-tuned, the five-fold stratified cross-validation regime was maintained during fine-tuning, and we ensured that the same testing split from fine-tuning was used during training of the gradient boosting classifier.

### G. SOFTWARE AND COMPUTING MACHINE

All experiments were conducted on the same computing machine equipped with a 16-core CPU, 64 GB RAM, and Titan X 24 GB RAM GPU. The software needed for the experiments was written in Python3 with the use of publicly available packages such as Flair [6], XGBoost, Transformers, Scikit-learn, and others.

## III. RESULTS AND DISCUSSION

### A. PARAMETER OPTIMIZATION OF TEXT GUIDE

In total, the design of the parameter optimization process proposed in Section 2.2 for three datasets required the computation of 600 test runs with pre-trained models. Table 4 presents summarized results in this regard, whereas detailed results of this research are presented in Figures 5–7, from which we can draw general conclusions. The most important conclusion is that parameter optimization has a significant impact on *Text Guide* performance. This observation holds for all analyzed data sets and models, with the greatest differences observed for the DMOZ 4090+ data set and the Longformer model [10].

A comparison of Figures 5 and 6 indicates that some similar patterns can be observed for the Albert model with an NTA of 512 tokens deployed on the DMOZ 510-1500 and DMOZ 1500+ data sets. First, low values of 2–3 for the TN number are preferred to construct token groups by *Text Guide*. Second, regarding the size of the ending part of the original text instance used by *Text Guide*, it seems that a smaller size is better; however, the differences between the best-performing models were marginal. Third, regarding the size of the beginning part of the original text instance used by *Text Guide*, the best-performing test runs benefited from shorter beginning parts.

However, the above conclusions are not universal, as demonstrated by Figure 7, which presents results for the DMOZ 4090+ data set and the Longformer model. In this setup, using a higher TN number to construct token groups allowed *Text Guide* to achieving visibly better results. Moreover, the size of the beginning and ending part of the original text instance did not have an obvious effect on the Longformer model.

The different conclusions obtained from analyses performed with different models and different data sets indicate that the best performance of *Text Guide* can be achieved by conducting parameter optimization for the model and data set to be analyzed. Presumably, more experiments on different data sets and models are required to discover more universal truths regarding the optimal parameters for *Text Guide*.

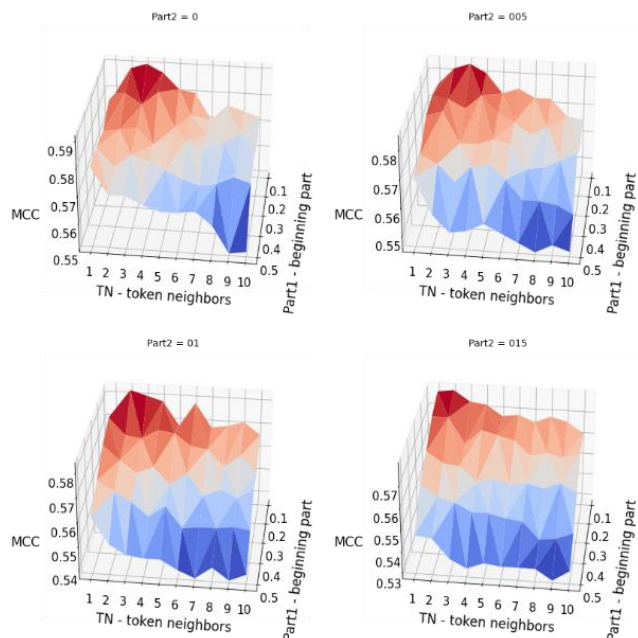

**FIGURE 6.** MCC of the pre-trained Albert base-v2 model on the DMOZ 1500+ data set. The title Part2 = 0 indicates that the ending part of the text was not explicitly considered, whereas greater Part2 values indicate that a greater number of tokens from the end of the original text instance were considered.

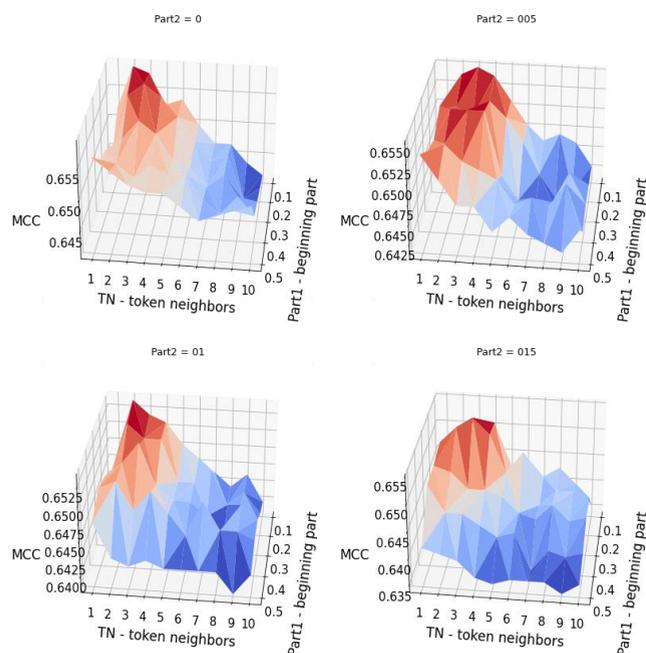

**FIGURE 5.** MCC of the pre-trained Albert base-v2 model on the DMOZ 510-1500 data set. The title Part2 = 0 indicates that the ending part of the text was not explicitly considered, whereas greater Part2 values indicate that a greater number of tokens from the end of the original text instance were considered.

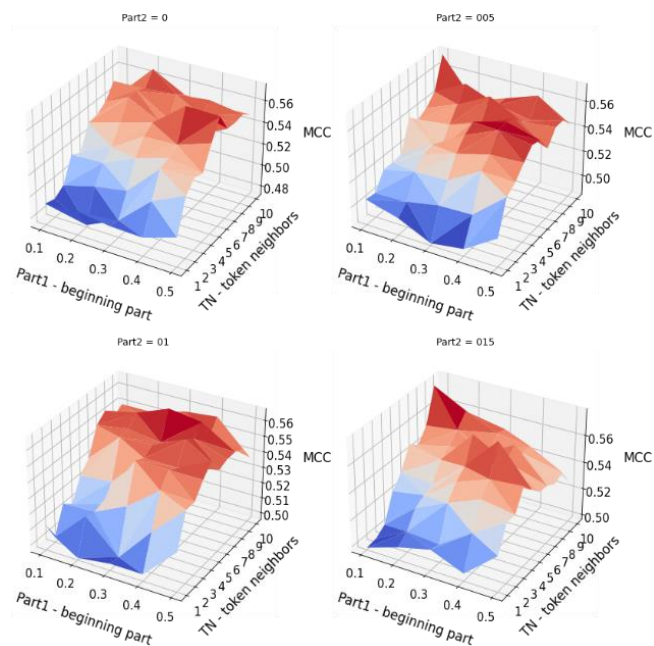

**FIGURE 7.** MCC of the pre-trained Longformer base model on the DMOZ 4090+ data set. The title Part2 = 0 indicates that the ending part of the text was not explicitly considered, whereas greater Part2 values indicate that a greater number of tokens from the end of the original text instance were considered.

TABLE IV
PERFORMANCE COMPARISON OF SELECTED PRE-TRAINED MODELS. 'X' INDICATES THAT THE GIVEN CONFIGURATION WAS NOT TESTED. IN OUR STUDY, FOR THE 'PART1 HEAD + PART2 TAIL' METHOD, WE ADOPTED PART1 = 0.75 AND PART2 = 0.25.

|  |  | Data Set | | |
|---|---|---|---|---|
| Model | Truncation Method | DMOZ 510-1500 | DMOZ 1500+ | DMOZ 4090+ |
| Albert Base-v2 | Naive Head-only | 0.6379 | 0.5359 | 0.44 |
| Albert Base-v2 | Part1 Head + Part2 Tail | 0.6396 | 0.5378 | 0.4512 |
| Albert Base-v2 | Text Guide | 0.659 | 0.5913 | 0.5138 |
| Longformer Base | Naive Head-only | 0.7162 | 0.5818 | 0.4877 |
| Longformer Base | Part1 Head + Part2 Tail | X | X | 0.489 |
| Longformer Base | Text Guide | X | X | 0.5784 |

## B. TEXT GUIDE AND COMPARISON OF FINE-TUNED MODELS

Interestingly, the positive effect of utilizing *Text Guide* is more evident for pre-trained than for fine-tuned models. When the former is analyzed, discussing the performance of *Text Guide* is simple, as *Text Guide* provided a great improvement in all analyzed trials. However, when the latter is considered, the conclusions are not as clear.

### 1) INITIAL OBSERVATIONS

Our initial examination of experiments with fine-tuned models led to several observations, which are primarily based on Table 5 and Figure 4. 1) For DMOZ 510-1500, which has the shortest original text instances, Text Guide did not provide any improvement. 2) For the DMOZ 1500+ data set, Text Guide provided an evident improvement over the compared truncation methods. 3) For DMOZ 4090+, the Albert model with Text Guide provided only a minor improvement over the "Part1 Head + Part2 Tail" method. 4) For DMOZ 4090+, the Longformer model with Text Guide provided significant improvement over basic truncation methods. 5) Comparing the performance of a plain BoW model with Albert shows that the latter model is superior for the DMOZ 510-1500, no matter the truncation method. For DMOZ 1500+ Albert was better than BoW only owing to Text Guide, and for the longest text instances found in DMOZ 4090+ BoW was superior. 6) For all data sets, the Roberta base model with Text Guide provided visibly better performance compared to other models.

TABLE V.
PERFORMANCE COMPARISON OF FINE-TUNED MODELS. 'X' INDICATES THAT THE MODEL WAS NOT TRAINED IN THE GIVEN CONFIGURATION. TEXT GUIDE RESULTS FOR THE DMOZ 510-1500 AND DMOZ 1500+ DATA SETS WERE OBTAINED AFTER PARAMETER OPTIMIZATION WAS PERFORMED ON PRE-TRAINED MODELS FOR EACH DATA SET SEPARATELY. THE TEXT GUIDE RESULT INDICATED BY "*" WAS OBTAINED AFTER ADDITIONAL TEXT GUIDE MODIFICATION. THE TEXT GUIDE RESULT INDICATED BY "**" WAS OBTAINED WITHOUT A DATA SET-SPECIFIC PARAMETER SEARCH; INSTEAD, TEXT GUIDE PARAMETERS WERE BORROWED FROM THE OPTIMIZATION PROCESS FOR THE DMOZ 1500+ DATA SET. THE TEXT GUIDE RESULT INDICATED BY "***" WAS OBTAINED AFTER ALLOWING MORE FINE-TUNING EPOCHS AND ADDITIONAL TEXT GUIDE MODIFICATION. THE MODIFICATIONS ARE DISCUSSED IN SECTION III.B.

|  |  | Data Set | | |
|---|---|---|---|---|
| Model | Truncation Method | DMOZ 510-1500 | DMOZ 1500+ | DMOZ 4090+ |
| Albert Base-v2 | Naive Head-only | 0.8958 | 0.8312 | 0.7589 |
| Albert Base-v2 | Part1 Head + Part2 Tail | 0.8966 | 0.8283 | 0.7719 |
| Albert Base-v2 | Text Guide | 0.8917 and 0.9001* | 0.8441 | 0.7799** |
| Longformer Base | Naive Head-only | X | X | 0.7316 |
| Longformer Base | Part1 Head + Part2 Tail | X | X | 0.7437 |
| Longformer Base | Text Guide | X | X | 0.7776 and 0.7746*** |
| Plain BoW | Whole text Instance | 0.8904 | 0.8355 | 0.7985 |
| RoBERTa Base | Text Guide | 0.9258 | 0.883 | 0.8554 |

### 2) DETAILED ANALYSIS AND FOLLOW-UP EXPERIMENTS

After deriving initial conclusions based on fine-tuning experiments, we further investigated observation 1 in conjunction with observation 2, as it seemed that *Text Guide* allows for improvement when the length of the original text instances exceeds the NTA model limit to a greater extent. In this regard, Figure 4 helped us to confirm that approximately half of the text instances in DMOZ 510-1500 exceeded the NTA model limit by a factor of only 0.5. Therefore, we conducted a small experiment with additional *Text Guide* modification, in which we utilized the naive head-only truncation method for original text instances that did not exceed the length of 1.5*NTA limit = 765 tokens while utilizing *Text Guide* for longer text instances. The result obtained with this modification, indicated by "*" in Table 5, provided an improvement over the original *Text Guide* result, confirming that *Text Guide* may be beneficial, especially if the length of the original text instance exceeds the NTA model limit to a large extent.

Next, we investigated preliminary observation 3. Initially, this conclusion seemed contradictory, as the above conclusions indicate that the improvement of a fine-tuned Albert model with *Text Guide* for DMOZ 4090+ should be much greater than that for DMOZ 1500+. Instead, the analyzed method outperformed known truncation methods by only a small margin. To explain this phenomenon, we note that the result of the Albert model with *Text Guide* for DMOZ 4090+ was obtained without data set- or model-specific parameter optimization in this study. Instead, the *Text Guide* parameters were borrowed from DMOZ 1500+. Therefore, it is possible that parameter optimization for this data set could further improve the results of this experiment.

Further, we concluded that a discussion of observation 4 should extend beyond simply noting the success of *Text Guide*. Of course, analyzing this set of fine-tuned Longformer results indicates that *Text Guide* allowed for a great improvement in performance; however, this conclusion is undermined by the performance comparison of the Albert and Longformer models for DMOZ 4090+, which does not appear as expected. The fact that the Albert model obtained better results than the Longformer model for extremely long text instances raises a question regarding the quality of the fine-tuning procedure of the Longformer model. If the fine-tuning procedure was faulty, the conclusion regarding the benefit of *Text Guide* in this regard may be biased. We first note differences between the adopted fine-tuning procedures for the Albert and Longformer models. The first substantial difference in the fine-tuning procedures is the batch size. For the former model, the batch size was set to 24, whereas the batch size was 2 for the latter, as a result of the GPU RAM limitation of our computing machine. Unfortunately, we did not have sufficient resources to experiment with larger batch sizes for the Longformer model, as this would require very high values of GPU RAM. Therefore, in our study, we focused on other differences, i.e., the learning rate and epoch parameters, and conducted additional fine-tuning procedures for the Longformer model on the DMOZ 4090+ data set. Because the learning rate for the Longformer model was set to an order of magnitude lower than that of the Albert model in order for the model to converge, we allowed the model to spend more time on fine-tuning and increased the maximum number of epochs from 6 to 10. In addition, because our *Text Guide* modification tested for the Albert model on the DMOZ 510-1500 data set resulted in enhanced performance, we applied the same modification of the *Text Guide* for Longformer and the DMOZ 4090+ data set. Eventually, for this data set length, approximately 40% of text instances were less than the 1.5*NTA Longformer limit = 6135 tokens. By performing a single experiment that required 18.2 hours of computation time, we obtained an MCC value of 0.7746, which is similar to the previous value for *Text Guide* with the Longformer model on the DMOZ 4090+ data set.

The above example shows that, unfortunately, experiments with large models in the fine-tuning regime are time-consuming and do not always provide the expected results. Additionally, due to the common problem of limited resources, practitioners and researchers are not always able to conduct thorough parameter optimization for large language models such as Longformer. In light of these considerations, the demonstrated performance of *Text Guide* with the Albert model on the DMOZ 4090+ data set warrants further investigation. Given our resource limitations, by using *Text Guide*, the Albert model, i.e., a small transformer model that did not require extensive optimization of fine-tuning parameters, slightly outperformed the Longformer model. We believe this to be an important conclusion from the perspective of real-life and limited resources.

Finally, a detailed analysis of the results of plain BoW models briefly discussed in observation 5 allows us to state that this type of models can be treated as a solid baseline in long text classification. Specifically, when dealing with data sets containing text instances that exceed the NTA limit only to a small extent, BoW models are likely to be outperformed even by small transformer models. However, the longer the text instances, the more difficult it becomes for transformer models to beat the performance of BoW models. As observed in point 6, using proper truncation methods along with larger state-of-the-art transformer models, such as Roberta base, allows outperforming BoW models in all analyzed data sets and by a large margin.

### C. GENERAL CONCLUSIONS

As presented in Tables 4 and 5, the use of *Text Guide* enables performance improvements over previous truncation methods, where the extent of improvement depends on the analyzed data set and adopted model. Thus, researchers and practitioners seeking to perform analyses with low computation costs can still benefit from standard transformer models with an NTA of 512 while improving their results through a new, non-costly truncation approach.

Moreover, as demonstrated for the DMOZ 4090+ data set, when the original text instance length exceeds the NTA limit of recent models specifically designed to handle long textual data, improvements achieved by *Text Guide* maybe even more meaningful. We include this statement because recent transformer models designed to analyze long text instances still cannot analyze whole extremely long instances and to cope with this fact, they simply truncate the original text instance. Therefore, even though they achieve state-of-the-research performance, they still lose information stored in the original text instances due to their use of naive or semi-naive truncation methods.

### D. CALL FOR FUTURE EXPERIMENTS

The method proposed in our study opens numerous areas for further investigation. First, *Text Guide* experiments on the DMOZ 4090+ data set with the Longformer model should be widened to other data sets with extremely long data instances, to other transformer model architectures created for analyses of such data, and to more detailed optimization of fine-tuning parameters. We believe that our preliminary results in this regard are promising and indicate that further research will prove the *Text Guide* useful for large transformer models and extremely long text instances. Unfortunately, experiments in this regard require high amounts of GPU RAM and weeks of computation time.

Second, our study indicated some discrepancy in conclusions derived from experiments conducted on pre-trained and fine-tuned models regarding the extent to which *Text Guide* improves results. An unexplored, promising, and highly resource-demanding research area that could bring more light in this regard is the possible benefit of optimizing *Text Guide* parameters based on fine-tuned models.

Third, one of our experiments with additionally modified *Text Guide*, which partially benefited from using naive head-only truncation for text instances slightly exceeding the NTA limit and using *Text Guide* for instances exceeding the limit, allowed for a slight increase in performance. The extent to which this approach can be leveraged should also be examined and threshold values other than the arbitrarily selected 1.5*NTA explored.

Fourth, *Text Guide* in phase 1 benefits from the XAI method embedded in the gradient boosting classifier. It may be useful to experiment with other classifiers and XAI methods, such as Shapley Additive exPlanations [25], for obtaining the sITFL.

Fifth, another possible extension of *Text Guide* could benefit not only from sorting important token features but from exploiting the actual values of feature importances, for example, by proposing a weighting algorithm. This approach could be explored while considering the fact that some important tokens appear more than once in the original text instance while *Text Guide* in its present form was designed to search only for the first occurrence of an important feature token. Changes in this regard could have a substantial impact on *Text Guide* performance.

Finally, *Text Guide* can be improved by the model used for identifying important feature tokens. We utilized BoW based on unigrams as an example in our study; thus, for improvements in this regard, future studies could focus on an extension to ngrams.

Unfortunately, pursuing all of the proposed lines of research is far beyond our resources. As demonstrated by the fine-tuning issues observed for the Longformer model, such efforts require not only substantial computational time but also more powerful computing machines capable of handling more than two text instances at a time during fine tuning of a large model.

We hope that our work will attract other researchers to further study the proposed method owing to the performance benefits, simplicity, and low computational cost of applying *Text Guide* once parameter optimization has been completed.

## IV. CONCLUSIONS

### A. GENERAL CHARACTERISTICS OF TEXT GUIDE

*Text Guide* is computationally inexpensive. As a preprocessing step, one must obtain a data set-level sITFL. In terms of computational cost per text instance, when compared with the other methods presented in Table 1, we believe that *Text Guide* is slightly more computationally expensive than the simplest truncation methods. Subjectively, we believe that *Text Guide* should receive a computational cost score of 2 in terms of Table 1, as it is much less expensive than the methods with higher scores.

As a truncation method, *Text Guide* loses information, but not as much information as simpler truncation methods. This claim is justified by the results demonstrated in this study. The performance of *Text Guide* strongly depends on the parameters utilized in phase 2, and the best results are obtained if the parameters are optimized. The extent to which parameter optimization can influence the performance of *Text Guide* is significant, as demonstrated in this work.

### B. FINAL REMARKS

In this study, we briefly reviewed existing approaches for addressing the classification of textual instances that exceed the length limits of the best-performing transformer models. We proposed a new method called *Text Guide* that can improve model classification performance when the length of the analyzed text instance exceeds the model limit. Compared with existing approaches based on either semi-naive truncation methods or very sophisticated deep learning approaches, *Text Guide* is an intermediate option, i.e., it applies a truncation algorithm guided by tokens identified as important with known methods from the XAI domain. As a key advantage, this approach does not interfere with language models selected for the final text classification process and provides good results for texts exceeding the model limit. Notably, *Text Guide* can also be used to improve the performance of recent transformer models specifically designed for the analysis of very long textual data if their increased analysis length limit is still exceeded by the text instance length.

## REFERENCES

[1] T. Mikolov, I. Sutskever, K. Chen, G. Corrado, and J. Dean, "Distributed Representations of Words and Phrases and their Compositionality," Oct. 2013, Accessed: Apr. 18, 2021. [Online]. Available: https://arxiv.org/abs/1310.4546v1

[2] L. Huang, D. Ma, S. Li, X. Zhang, and H. WANG, "Text Level Graph Neural Network for Text Classification," *arXiv:1910.02356 [cs]*, Oct. 2019, Accessed: Apr. 18, 2021. [Online]. Available: http://arxiv.org/abs/1910.02356

[3] A. Vaswani *et al.*, "Attention Is All You Need," *arXiv:1706.03762 [cs]*, Dec. 2017, Accessed: Jan. 16, 2021. [Online]. Available: http://arxiv.org/abs/1706.03762

[4] J. Pennebaker, M. Francis, and R. Booth, "Linguistic inquiry and word count (LIWC)," Jan. 2001.

[5] S. A. Crossley, K. Kyle, and D. S. McNamara, "Sentiment Analysis and Social Cognition Engine (SEANCE): An automatic tool for sentiment, social cognition, and social-order analysis," *Behav Res*, vol. 49, no. 3, pp. 803–821, Jun. 2017, doi: 10.3758/s13428-016-0743-z.

[6] A. Akbik, T. Bergmann, D. Blythe, K. Rasul, S. Schweter, and R. Vollgraf, "FLAIR: An Easy-to-Use Framework for State-of-the-Art NLP," in *Proceedings of the 2019 Conference of the North American Chapter of the Association for Computational Linguistics (Demonstrations)*, Minneapolis, Minnesota, Jun. 2019, pp. 54–59. doi: 10.18653/v1/N19-4010.

[7] C. Sun, X. Qiu, Y. Xu, and X. Huang, "How to Fine-Tune BERT for Text Classification?," *arXiv:1905.05583 [cs]*, Feb. 2020, Accessed: Apr. 18, 2021. [Online]. Available: http://arxiv.org/abs/1905.05583

[8] A. Mulyar, E. Schumacher, M. Rouhizadeh, and M. Dredze, "Phenotyping of Clinical Notes with Improved Document Classification Models Using Contextualized Neural Language Models," *arXiv:1910.13664 [cs]*, Sep. 2020, Accessed: Apr. 18, 2021. [Online]. Available: http://arxiv.org/abs/1910.13664

[9] R. Zhang, Z. Wei, Y. Shi, and Y. Chen, "BERT-AL: BERT for Arbitrarily Long Document Understanding," Sep. 2019, Accessed: Apr. 18, 2021. [Online]. Available: https://openreview.net/forum?id=SklnVAEFDB

[10] I. Beltagy, M. E. Peters, and A. Cohan, "Longformer: The Long-Document Transformer," *arXiv:2004.05150 [cs]*, Dec. 2020, Accessed: Apr. 18, 2021. [Online]. Available: http://arxiv.org/abs/2004.05150

[11] K. Choromanski *et al.*, "Rethinking Attention with Performers," *arXiv:2009.14794 [cs, stat]*, Mar. 2021, Accessed: Jun. 13, 2021. [Online]. Available: http://arxiv.org/abs/2009.14794

[12] N. Kitaev, Ł. Kaiser, and A. Levskaya, "Reformer: The Efficient Transformer," *arXiv:2001.04451 [cs, stat]*, Feb. 2020, Accessed: Jun. 13, 2021. [Online]. Available: http://arxiv.org/abs/2001.04451

[13] S. Min, V. Zhong, R. Socher, and C. Xiong, "Efficient and Robust Question Answering from Minimal Context over Documents," *arXiv:1805.08092 [cs]*, May 2018, Accessed: Apr. 18, 2021. [Online]. Available: http://arxiv.org/abs/1805.08092

[14] M. Ding, H. Yang, C. Zhou, and J. Tang, "CogLTX: Applying BERT to Long Texts," p. 13.

[15] M. Zaheer *et al.*, "Big Bird: Transformers for Longer Sequences," *arXiv:2007.14062 [cs, stat]*, Jan. 2021, Accessed: Apr. 18, 2021. [Online]. Available: http://arxiv.org/abs/2007.14062

[16] Y. Tay, M. Dehghani, D. Bahri, and D. Metzler, "Efficient Transformers: A Survey," *arXiv:2009.06732 [cs]*, Sep. 2020, Accessed: Apr. 18, 2021. [Online]. Available: http://arxiv.org/abs/2009.06732

[17] Y. Tay *et al.*, "Long Range Arena: A Benchmark for Efficient Transformers," Nov. 2020, Accessed: Apr. 18, 2021. [Online]. Available: https://arxiv.org/abs/2011.04006v1

[18] F. Krzysztof, *TextGuide*. 2021. Accessed: Jun. 13, 2021. [Online]. Available: https://github.com/krzysztoffiok/TextGuide

[19] "Transformers." index.html (accessed Apr. 18, 2021).

[20] "Home Page for 20 Newsgroups Data Set." http://qwone.com/~jason/20Newsgroups/ (accessed Apr. 18, 2021).

[21] G. Sood, "Parsed DMOZ data." Harvard Dataverse, Feb. 18, 2021. doi: 10.7910/DVN/OMV93V.

[22] Z. Lan, M. Chen, S. Goodman, K. Gimpel, P. Sharma, and R. Soricut, "ALBERT: A Lite BERT for Self-supervised Learning of Language Representations," *arXiv:1909.11942 [cs]*, Feb. 2020, Accessed: Jan. 16, 2021. [Online]. Available: http://arxiv.org/abs/1909.11942

[23] Y. Liu *et al.*, "RoBERTa: A Robustly Optimized BERT Pretraining Approach," *arXiv:1907.11692 [cs]*, Jul. 2019, Accessed: Jan. 16, 2021. [Online]. Available: http://arxiv.org/abs/1907.11692

[24] D. Chicco and G. Jurman, "The advantages of the Matthews correlation coefficient (MCC) over F1 score and accuracy in binary classification evaluation," *BMC Genomics*, vol. 21, no. 1, p. 6, Jan. 2020, doi: 10.1186/s12864-019-6413-7.

[25] S. Lundberg and S.-I. Lee, "A Unified Approach to Interpreting Model Predictions," *arXiv:1705.07874 [cs, stat]*, Nov. 2017, Accessed: Apr. 18, 2021. [Online]. Available: http://arxiv.org/abs/1705.07874

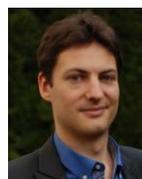

**KRZYSZTOF FIOK** received his Master of Science and Engineer in Transportation diploma in 2009 at Warsaw University of Technology, Poland. In 2015 he received his PhD at the same University. Since 2019 he is a Postdoc at the Faculty of Industrial Engineering and Management Systems, University of Central Florida, Orlando. His research interests are mostly focused on artificial intelligence and machine learning applied to natural language, image analysis, biomedical signals, and other domains. He also has experience in ergonomics.

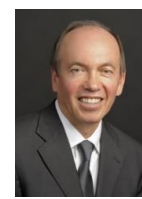

**WALDEMAR KARWOWSKI** (Senior Member, IEEE) received the M.S. degree in production engineering and management from the Technical University of Wroclaw, Poland, in 1978, and the Ph.D. degree in industrial engineering from Texas Tech University, in 1982. He is currently the Pegasus

Professor and the Chairman of the Department of Industrial Engineering and Management Systems and the Executive Director of the Institute for Advanced Systems Engineering, University of Central Florida, Orlando, FL, USA. His publications focus on mathematical modeling and computer simulation with applications to human systems engineering, human-centered-design, safety, neuro-fuzzy systems, nonlinear dynamics and chaos, and neuroergonomics. He serves as the Co-Editor-in-Chief for the Journal Theoretical Issues in Ergonomics Science (Taylor and Francis, Ltd.), the Editor-in-Chief for the Human-Intelligent Systems Integration Journal (Springer), and the Field Chief Editor of the Frontiers in Neuroergonomics Journal.

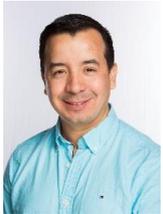

**EDGAR GUTIERREZ-FRANCO** his educational background includes a B.S. in Industrial Engineering from the University of La Sabana (2004, Colombia). MSc. in Industrial Engineering, from the University of Los Andes (2008, Colombia) and a Ph.D. in Industrial Engineering and Management Systems from the University of Central Florida (2019, USA). His expertise includes the use of machine learning, operation research, and simulation techniques for supply chain management and systems modeling. Dr. Edgar is Research Affiliate at the Center for Latin-American Logistics Innovation (CLI) part of the MIT Global SCALE Network and a Postdoctoral Associate at the Massachusetts Institute of Technology. He has experience in consultancy, retail, and beverage industry. During his stay at the Center for Transportation and Logistics, Massachusetts Institute of Technology (2009-2010) he participated in. projects of MIT CTL's in Supply Chain Innovation in Emerging Markets (city logistics) and Carbon-Efficient Supply Chains/Sustainability.

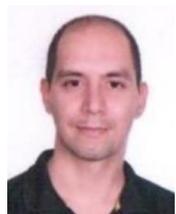

**MOHAMMAD REZA DAVAHLI** received the B.S. degree in civil engineering from Isfahan University of Technology, Iran. He worked five years as a data analyst in Kayson Company, which is one of the leading engineering, procurement, and construction companies. He is currently pursuing the Ph.D. degree with the Industrial Engineering and Management Systems Program, University of Central Florida. At present, he is a Research Assistant, advised by Prof. Waldemar Karwowski. His current research interests include time-series data analysis, natural language processing, and sequence-learning predictive models.

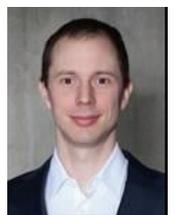

**MACIEJ WILAMOWSKI** was born in Warsaw, Poland in 1985. He received M.S and PhD degrees in economics from the University of Warsaw in 2010 and 2014 respectively. From 2015 he is an Assistant Professor at University of Warsaw. His research interests include experimental economics, scientometrics and machine learning. He is the author of more than 10 articles in these fields. In 2017 he founded a boutique consulting firm that specializes in development and deployment of machine learning in business environments.

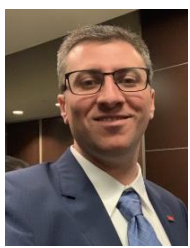

**TAREQ AHRAM**, PhD, is the Lead Scientist and Assistant Research Professor working at the Institute for Advanced Systems Engineering (IASE) at the University of Central Florida. He holds Ph.D. (2008) in Industrial Engineering from the University of Central Florida with specialization in Human Systems Integration and large-scale information retrieval systems optimization (database optimization algorithms). Dr. Ahram served as an invited speaker and scientific member at several Systems Engineering, Emerging Technologies, Neurodesign and Human Factors research and as invited speaker and program committee member at the Department of Defense Human Systems Integration and Human Factors Engineering Technical Advisory Group.

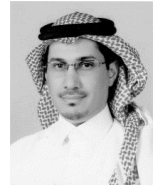

**AWAD ALJUAID** is an Associate Professor at Industrial Engineering Department at Taif University TU, Saudi Arabia. He received the B.S.(2003) in Systems Engineering from King Fahd University of Petroleum and Minerals, Saudi Arabia. Also, he received the M.S. (2009) in Industrial Engineering from King Abdul-Aziz University, Saudi Arabia, and a Ph.D. (2016) in Industrial Engineering from University of Central Florida, USA. He has been working as safety engineer at SABIC (2003) and planning engineer at SCECO for 7 years. He is a former Dean of University Development Deanship at TU and Director of Strategic Planning and Information Department at TU . He has several publications focused on, neuroergonomic, safety, quality assurance, planning , neuro-fuzzy system, and supply chain management. He serves as editor for the Journal Theoretical Issues in Ergonomics Science (Taylor and Francis, Ltd.) and he is reviewer and consultant for institutional accreditation in Saudi Arabia.

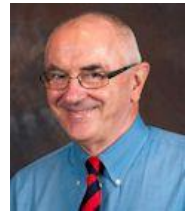

**JOZEF ZURADA** was born in Sosnowiec, Poland, in 1949. He received the M.S. degree in electrical engineering from the Technical University of Gdansk, Poland, in 1972; the Ph.D. degree in computer science engineering from the University of Louisville in 1995; and the D.Sc. degree in technical informatics from the Systems Research Institute of the Polish Academy of Sciences in 2014. Since 2005 he has been a Professor in the Information Systems, Analytics, and Operations Department at the University of Louisville. He was a visiting scholar at the School of Computer and Information Science, Edith Cowan, University, Perth, Australia, in 2007; and the Department of Computer and Electrical Engineering, University of Alberta, Edmonton, Canada, in 2007. His research interests include applications of advanced computational intelligence and soft computing methods for assisting in decision making in business and manufacturing systems as well as streaming data analytics. He published about 110 refereed articles and delivered over 60 presentations.